\documentclass[10pt,twocolumn,letterpaper]{article}

\usepackage[pagenumbers]{cvpr} 
\usepackage{graphicx}
\usepackage{amsmath}
\usepackage{amssymb}
\usepackage{booktabs}
\usepackage[pagebackref=true,breaklinks=true,colorlinks,bookmarks=false]{hyperref}
\usepackage{color,xcolor}
\usepackage{epsfig}

\usepackage{adjustbox}
\usepackage{array}
\usepackage{booktabs}
\usepackage{colortbl}
\usepackage{float,wrapfig}
\usepackage{hhline}
\usepackage{multirow}
\usepackage{dsfont}

\usepackage{amsmath,amsfonts,amsthm,amssymb}
\usepackage{bm}
\usepackage{nicefrac}
\usepackage{microtype}

\usepackage{changepage}
\usepackage{extramarks}
\usepackage{fancyhdr}
\usepackage{lastpage}
\usepackage{setspace}
\usepackage{soul}
\usepackage{xspace}
\usepackage{pifont}

\usepackage{algorithm, algorithmic}
\usepackage{enumerate}
\usepackage{listings}
\newcolumntype{L}[1]{>{\raggedright\let\newline\\\arraybackslash\hspace{0pt}}m{#1}}
\newcolumntype{C}[1]{>{\centering\let\newline\\\arraybackslash\hspace{0pt}}m{#1}}
\newcolumntype{R}[1]{>{\raggedleft\let\newline\\\arraybackslash\hspace{0pt}}m{#1}}

\newcommand{\fig}[1]{Figure~\ref{#1}}

\newcommand{\tab}[1]{Table~\ref{#1}}

\newcommand{\ignorethis}[1]{}

\makeatletter
\DeclareRobustCommand\onedot{\futurelet\@let@token\@onedot}
\def\@onedot{\ifx\@let@token.\else.\null\fi\xspace}

\def\ie{\emph{i.e}\onedot}

\makeatother

\definecolor{citecolor}{RGB}{34,139,34}
\definecolor{mydarkblue}{rgb}{0,0.08,1}
\definecolor{mydarkgreen}{rgb}{0.02,0.6,0.02}
\definecolor{mydarkred}{rgb}{0.8,0.02,0.02}
\definecolor{mydarkorange}{rgb}{0.40,0.2,0.02}
\definecolor{mypurple}{RGB}{111,0,255}
\definecolor{myred}{rgb}{1.0,0.0,0.0}
\definecolor{mygold}{rgb}{0.75,0.6,0.12}
\definecolor{myblue}{rgb}{0,0.2,0.8}
\definecolor{mydarkgray}{rgb}{0.66,0.66,0.66}

\usepackage[capitalize]{cleveref}
\crefname{section}{Sec.}{Secs.}
\Crefname{section}{Section}{Sections}
\Crefname{table}{Table}{Tables}
\crefname{table}{Tab.}{Tabs.}

\def\cvprPaperID{5679}
\def\confName{CVPR}
\def\confYear{2022}

\begin{document}
\title{Sparse Fuse Dense: Towards High Quality 3D Detection with Depth Completion}
\author{Xiaopei Wu\textsuperscript{\rm 1,2}, Liang Peng\textsuperscript{\rm 1,2}, Honghui Yang\textsuperscript{\rm 1,2}, Liang Xie\textsuperscript{\rm 1}, Chenxi Huang\textsuperscript{\rm 1},\\ Chengqi Deng\textsuperscript{\rm 1}, Haifeng Liu\textsuperscript{\rm 1}, Deng Cai\textsuperscript{\rm 1,2\thanks{Corresponding author}*} \\
	\textsuperscript{\rm 1}State Key Lab of CAD\&CG, Zhejiang University \quad
	\textsuperscript{\rm 2}Fabu Inc., Hangzhou, China \\
	{\tt\small \{wuxiaopei, pengliang, yanghonghui\}@zju.edu.cn}}

\maketitle
\vspace*{-5mm}
\begin{abstract}
	Current LiDAR-only 3D detection methods inevitably suffer from the sparsity of point clouds. 
	Many multi-modal methods are proposed to alleviate this issue, while different representations of images and point clouds make it difficult to fuse them, resulting in suboptimal performance. 
	In this paper, we present a novel multi-modal framework \textbf{SFD} (Sparse Fuse Dense), which utilizes pseudo point clouds generated from depth completion to tackle the issues mentioned above. Different from prior works, we propose a new RoI fusion strategy \textbf{3D-GAF} (3D Grid-wise Attentive Fusion) to make fuller use of information from different types of point clouds. Specifically, 3D-GAF fuses 3D RoI features from the pair of point clouds in a grid-wise attentive way, which is more fine-grained and more precise.
	In addition, we propose a \textbf{SynAugment} (Synchronized Augmentation) to enable our multi-modal framework to utilize all data augmentation approaches tailored to LiDAR-only methods. Lastly, we customize an effective and efficient feature extractor \textbf{CPConv} (Color Point Convolution) for pseudo point clouds. It can explore 2D image features and 3D geometric features of pseudo point clouds simultaneously.
	Our method holds the highest entry on the KITTI car 3D object detection leaderboard\footnote[2]{On the date of CVPR deadline,~\ie, Nov.16, 2021}, demonstrating the effectiveness of our SFD. Codes are available at \url{https://github.com/LittlePey/SFD}.
\end{abstract}
\section{Introduction}
\label{sec:intro}

In recent years, the rise of deep learning and autonomous driving has led to a rapid development of 3D detection. Current 3D detection methods are mainly based on LiDAR point clouds \cite{shi2019pointrcnn,Chen2019fastpointrcnn,yang2019std,liu2020tanet,shi2020pv, Ye_2020_CVPR,zheng2021se,Chai_2021_CVPR,ge2021real,mao2021pyramid}, while the sparsity of point clouds considerably limits their performances. The sparse LiDAR point clouds provide poor information in distant and occluded regions, making it difficult to generate precise 3D bounding boxes. Many multi-modal methods are proposed to address this problem. MV3D \cite{MV3D} introduces an RoI fusion strategy to fuse features of images and point clouds on the second stage.
AVOD \cite{AVOD} proposes to fuse full resolution feature crops from the image feature maps and BEV feature maps for a high recall.
MMF \cite{liang2019multi} leverages 2D detection, ground estimation and depth completion to assist 3D detection. In MMF, pseudo point clouds are used for backbone feature fusion, and depth completion feature maps are used for RoI feature fusion. Despite their great success, they have two shortcomings.

\textit{Coarse RoI Fusion Strategy}\quad 
When fusing RoI features, as shown in \fig{fig:method_previous}(a), previous RoI fusion methods concatenate \textit{2D LiDAR RoI features} cropped from BEV LiDAR feature maps and \textit{2D image RoI features} cropped from FOV image feature maps. We note that this RoI fusion strategy is coarse. Firstly, 2D image RoI features are usually mixed with features from other objects or backgrounds, which will confuse the model. Secondly, the RoI fusion strategy ignores object part correspondences in 2D images and 3D point clouds. In this paper, we propose a more fine-grained RoI fusion strategy \textbf{3D-GAF} (3D Grid-wise Attentive Fusion), which fuse \textit{3D RoI features} instead of \textit{2D RoI features} as shown in \fig{fig:method_previous}(b). We elaborate on three advantages of 3D-GAF over previous RoI fusion methods in Section \ref{sec:3D_GAF}.

\textit{Insufficient Data Augmentation}\quad
This shortcoming exists in most multi-modal methods. Because 2D image data cannot be operated like 3D LiDAR data, many data augmentation approaches are difficult to deploy in multi-modal methods. It is a crucial reason why multi-modal methods are usually inferior to single-modal methods \cite{zhang2020multi}. To this end, we introduce our \textbf{SynAugment} (Synchronized Augmentation). We observe that after converting 2D images to 3D pseudo point clouds, the representations of images and raw point clouds are unified, suggesting that we can operate images just like raw point clouds. However, it is not enough. Some complicated data augmentation approaches such as gt-sampling \cite{yan2018second} and local rotation \cite{zheng2020cia} may cause occlusions on the FOV (field of view). It is a non-negligible issue because image features need to be extracted on the FOV. Now, it is time to jump out of the mindset. With 2D images converted to 3D pseudo point clouds, why don't we directly extract image features in 3D space? In this way, we no longer need to consider the FOV occlusion issue.

Nevertheless, it is non-trivial to extract features of pseudo point clouds in 3D space. Thus, we present a \textbf{CPConv} (Color Point Convolution), which searches neighbors of pseudo points on the image domain. It enables us to extract both image features and geometric features of pseudo point clouds efficiently. Considering the FOV occlusion issue, we cannot project all pseudo points to the image space of current frame for neighbor search. Here we propose an \textit{RoI-aware Neighbor Search}, which projects pseudo points in each 3D RoI to their original image space, as illustrated in \fig{fig:method_cpconv_augmentation}. Hence, pseudo points that occlude each other on the FOV will not become neighbors when performing neighbor search. In other words, their features will not interfere with each other.

To summarize, our contributions are listed as follows:
\begin{itemize}
	\item We propose a new RoI feature fusion strategy \textbf{3D-GAF} to fuse RoI features from raw point clouds and pseudo point clouds in a more fine-grained manner.
	\item We present a data augmentation method \textbf{SynAugment} to solve the insufficient data augmentation issue that multi-modal methods suffer from.
	\item We customize an effective and efficient feature extractor \textbf{CPConv} for pseudo point clouds. It can extract both 2D image features and 3D geometric features.
	\item We demonstrate the effectiveness of our method with extensive experiments. Specially, we rank $1^{st}$ on the KITTI car 3D object detection leaderboard.
\end{itemize}

\section{Related Work}
\label{sec:related}

\paragraph{3D Detection Using Single-modal Data.}
Current 3d detection methods are mainly based on LiDAR data. SECOND \cite{yan2018second} proposes a sparse convolution operation to speed up 3D convolution. SA-SSD \cite{he2020structure} exploits an auxiliary network to guide the features. PV-RCNN \cite{shi2020pv} leverages the advantages of voxel-based methods and point-based methods to get more discriminative features. Voxel-RCNN \cite{deng2020voxel} points out that precise positioning of raw points is unnecessary. SE-SSD \cite{zheng2021se} attains an excellent performance with self-ensembling. CenterPoint \cite{Yin_2021_CVPR} provides a simple but effective anchor-free framework for 3D detection. LiDAR R-CNN \cite{li2021lidar} gives an effective solution to remedy the scale ambiguity problem issue. SPG \cite{xu2021spg} generates semantic points to recover missing parts of the foreground objects. VoTr \cite{mao2021voxel} presents a transformer-based architecture to capture large context information efficiently. 
Pyramid R-CNN \cite{mao2021pyramid} designs a pyramid RoI head to adaptively learn the features from the sparse points of interest.
CT3D \cite{sheng2021improving} devises a channel-wise transformer to capture rich contextual dependencies
among points. However, LiDAR data is usually sparse, posing a challenge for these methods.
 
\paragraph{3D Detection Using Multi-modal Data.}
Due to the sparsity of point clouds, researchers seek help from multi-modal methods that utilize both images and point clouds. Some methods  \cite{qi2018frustum, xu2018pointfusion, wang2019frustum, zhao20193d} use a cascading approach to exploit multi-modal data. However, their performances are bounded by the 2D detector.
MV3D \cite{MV3D} realizes a two-stage multi-modal framework with an RoI feature fusion strategy that uses images for RoI refinement.
ContFuse \cite{CONTFUSE} proposes a continuous fusion layer to fuse BEV feature maps and image feature maps. 
MMF \cite{liang2019multi} benefits from multi-task learning and multi-sensor fusion.
VMVS \cite{ku2019improving} generates a set of virtual views for each detected pedestrian in pseudo point clouds. Then the different views are used to produce an accurate orientation estimation.
3D-CVF \cite{yoo20203d} fuses features from multi-view images. CLOCs PVCas \cite{pang2020clocs} refines confidences of 3D candidates with 2D candidates in a learnable manner.
Some works \cite{xie2020pi, vora2020pointpainting, huang2020epnet, sindagi2019mvx} realize a fine-grained fusion by establishing correspondence between images and point clouds, and then indexing image features by point clouds. However, the image information they index is limited due to the sparse correspondence between images and point clouds.
It is noteworthy that although MMF \cite{liang2019multi} also employs the depth completion, it does not solve the two issues mentioned in Section \ref{sec:intro}. In this paper, we make full use of pseudo point clouds and give an effective solution.

\vspace{-3mm}
\paragraph{Depth Completion.} Depth completion aims to predict a dense depth map from a sparse one with the guidance of a color image. Recently, many efficient depth completion methods are proposed \cite{hu2021penet, imran2021depth, gu2021denselidar, guizilini2021sparse}. 
\cite{hu2021penet} utilizes a two-branch backbone to realize a precise and efficient depth completion network. 
\cite{imran2021depth} proposes a multi-hypothesis depth representation that can sharp depth boundary between foreground and background.
Although the primary purpose of the depth completion task is to serve downstream tasks, there are few methods using depth completion in 3D detection. In the image-based 3D object detection, there are some works \cite{wang2019pseudo, you2019pseudo} that use depth estimation to generate pseudo point clouds. However, their performances are greatly limited due to the lack of accurate or sufficient raw LiDAR point clouds.

\section{Sparse Fuse Dense}

\subsection{Preliminaries}
For simplicity, we name the raw LiDAR point clouds generated by LiDAR and the pseudo point clouds generated from depth completion as \textit{raw clouds} and \textit{pseudo clouds}, respectively. Given a frame of raw clouds $\mathcal{R}$, we can convert it into a sparse depth map $\mathcal{S}$ with a known projection $T_{\rm{LiDAR}\rightarrow\rm{image}}$. Let $\mathcal{I}$ denote the image that corresponds to $\mathcal{R}$. Feeding $\mathcal{I}$ and $\mathcal{S}$ to a depth completion network, we can get a dense depth map $\mathcal{D}$. With a known projection $T_{\rm{image}\rightarrow\rm{LiDAR}}$, we can get a frame of pseudo clouds $\mathcal{P}$. 

\begin{figure*}[t]
	\centering
	\includegraphics[width=1\linewidth]{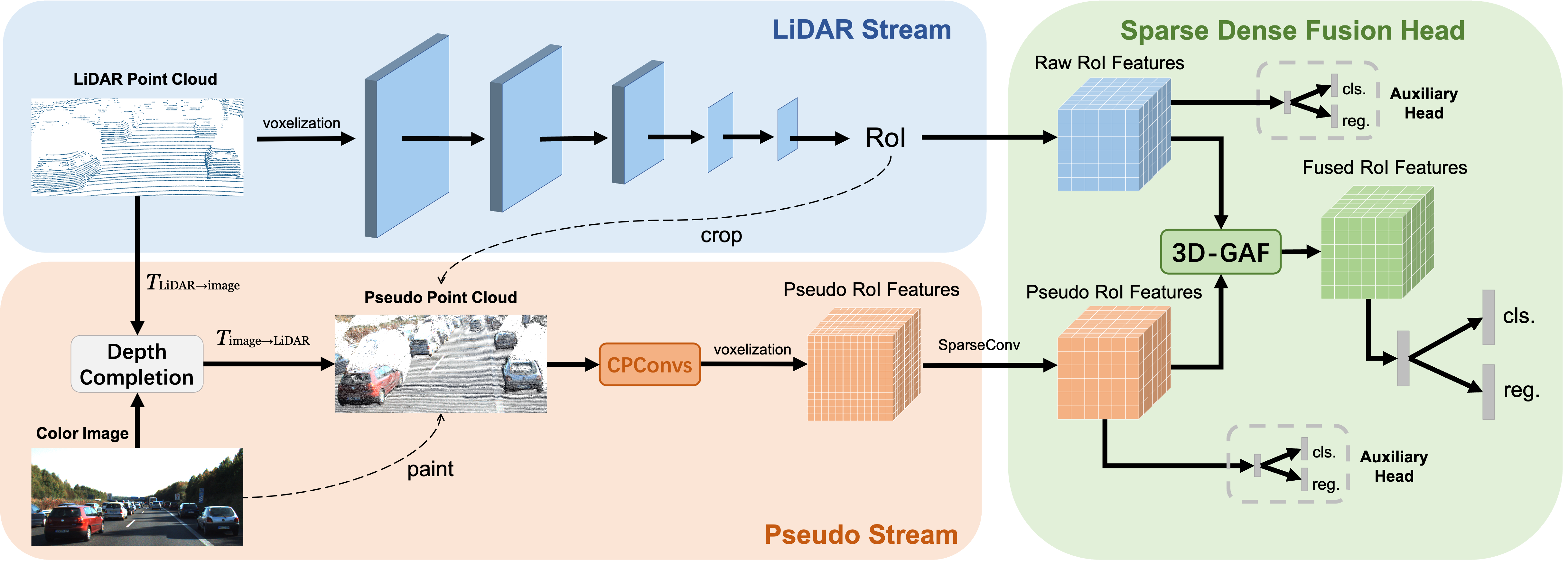}
	\caption{SFD consists of three parts: \textit{LiDAR Stream}, \textit{Pseudo Stream} and \textit{Sparse Dense Fusion Head}. \quad(1) \textit{LiDAR Stream} only uses raw clouds to predict 3D RoIs. Then RoIs are used to crop raw clouds and pseudo clouds. \quad(2) \textit{Pseudo Stream} uses raw clouds and images to generate pseudo clouds. Painting pseudo clouds with RGB, we get colorful pseudo clouds. Then several CPConvs (see Section \ref{sec:CPConv}) are performed to extract rich information of pseudo clouds in RoIs. At the end of Pseudo Stream, pseudo clouds in RoIs are voxelized, and 3D sparse convolutions are applied. \quad(3) In \textit{Sparse Dense Fusion Head}, RoI features from raw clouds and pseudo clouds are fused by 3D-GAF (see Section \ref{sec:3D_GAF}), then the fused RoI features are used to predict class confidences and bounding boxes. In addition, two auxiliary heads are employed to regularize our network. They can be detached at inference time.}
	\vspace{-10pt}
	\label{fig:method_overview}
\end{figure*}
\subsection{Overview of Methods}
We show our framework in \fig{fig:method_overview}, including: 
(1) a \textit{LiDAR Stream} using only raw clouds and serving as an RPN to produce 3D RoIs; 
(2) a \textit{Pseudo Stream} that extracts point features with proposed CPConv, and extracts voxel features with sparse convolutions; 
(3) a \textit{Sparse Dense Fusion Head} that fuses 3D RoI features from raw clouds and pseudo clouds in a grid-wise attentive manner, and produces final predictions. We detail our method in the following sections.

\subsection{3D Grid-wise Attentive Fusion}
\label{sec:3D_GAF}
\begin{figure}[t]
	\centering
	\includegraphics[width=0.82\linewidth]{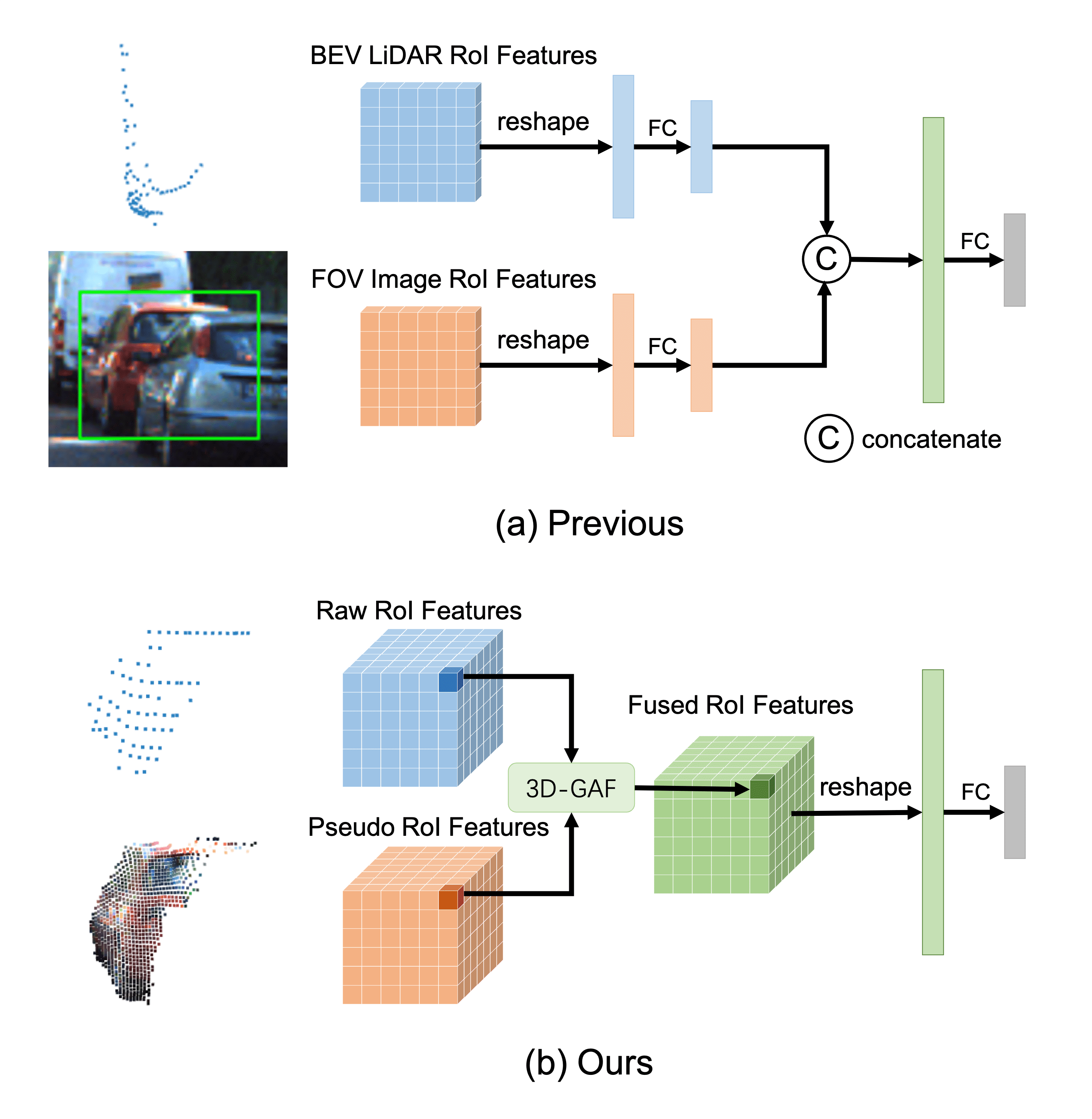}
	\vspace{-8pt}	
	\caption{Comparison between previous methods and 3D-GAF.}
	\vspace{-16pt}
	\label{fig:method_previous}
\end{figure}

Due to the dimensional gap between images and point clouds, previous works \cite{MV3D,AVOD, liang2019multi} directly concatenate \textit{2D LiDAR RoI features} cropped from BEV LiDAR feature maps and \textit{2D image RoI features} cropped from FOV image feature maps, which is a coarse RoI fusion strategy. In our method, with 2D images converted to 3D pseudo clouds, we can fuse the RoI features from images and point clouds in a more fine-grained manner, as shown in \fig{fig:method_previous}. Our 3D-GAF consists of 3D Fusion, Grid-wise Fusion and Attentive Fusion.

(1) \textit{\textbf{3D Fusion}}. We use a 3D RoI to crop 3D raw clouds and 3D pseudo clouds, which only includes LiDAR features and image features in the 3D RoI, as shown in \fig{fig:method_previous}(b). Previous methods use 2D RoI to crop image features, which will involve features from other objects or backgrounds. It causes a lot of interference, especially for occluded objects, as shown in \fig{fig:method_previous}(a).
(2) \textit{\textbf{Grid-wise Fusion}}. In previous RoI fusion methods, there are no correspondences between image RoI grids and LiDAR RoI grids, so they directly concatenate image RoI features and LiDAR RoI features. In our methods, thanks to the same representation of raw RoI features and pseudo RoI features, we can fuse each pair of grid features separately. It enables us to accurately enhance each part of an object with the corresponding pseudo grid features.
(3) \textit{\textbf{Attentive Fusion}}. Aiming to fuse each pair of grid features from raw RoI and pseudo RoI adaptively, we utilize a simple attention module motivated by \cite{hu2018squeeze, huang2020epnet, li2019selective}. Generally, we predict a pair of weights for each pair of grids and weight the pair of grid features with the weights to get the fused grid features. 
To validate the effectiveness of 3D Fusion, Grid-wise Fusion and Attentive Fusion, we provide ablation studies in Section \ref{sec:Experiments}.

\begin{figure*}[t]
	\begin{center}
		\setlength{\fboxrule}{0pt}
		\fbox{\includegraphics[width=0.98\textwidth]{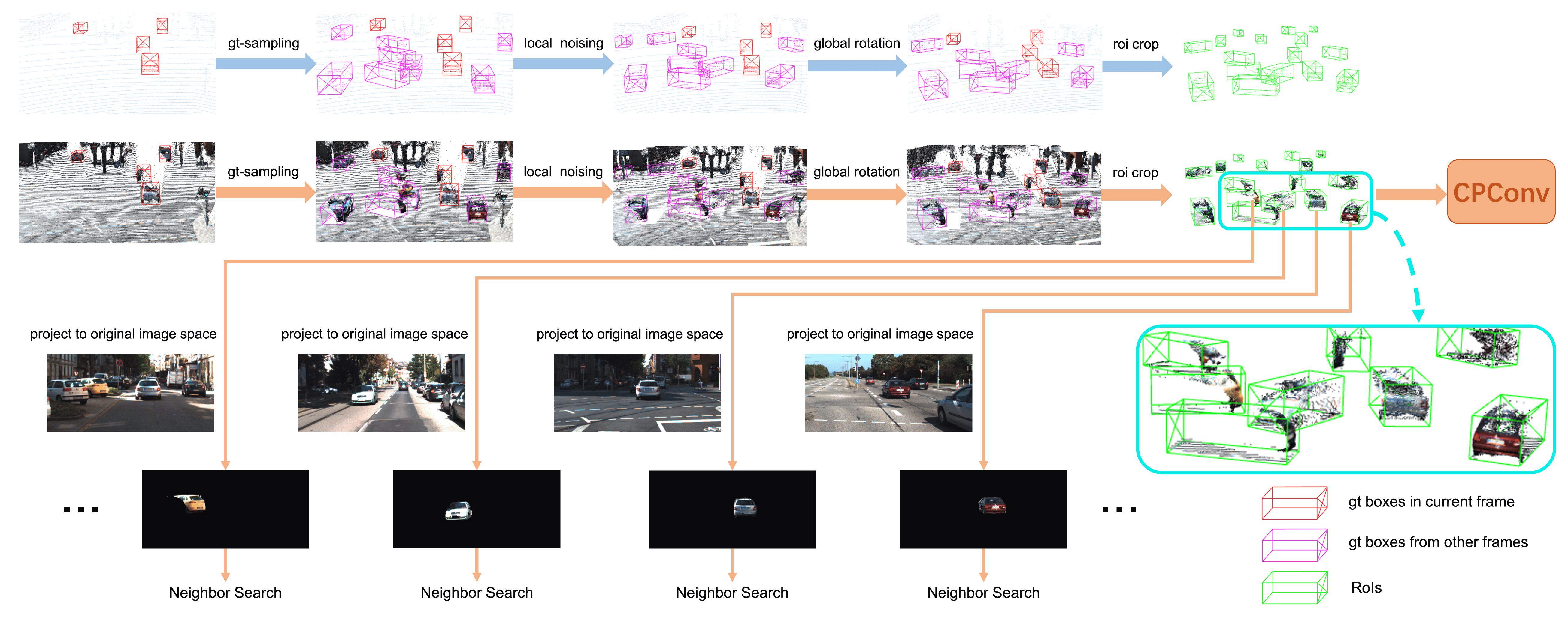}}
	\end{center}
	\captionsetup{font={small}}
	\vspace{-14pt}
	\caption{Illustration of SynAugment and RoI-aware Neighbor Search in CPConv. We show original gt boxes, sampled gt boxes and RoIs in red, purple and green, respectively. For the convenience of visualization, we only show 3 data augmentation approaches, and we remove some redundant and low score RoIs.}
	\label{fig:method_cpconv_augmentation}
	\vspace{-14pt}
\end{figure*}
Here we provide a detailed description of our 3D-GAF. Let $\mathbf{b}$ denote a single 3D RoI. We denote
$F^{\rm{raw}} \in \mathbb{R}^{n \times C}$ and $F^{\rm{pse}} \in \mathbb{R}^{n \times C}$ as the raw cloud RoI feature and pseudo cloud RoI feature in $\mathbf{b}$, respectively. Here $n$ ($6\times6\times6$ by default, following our baseline Voxel-RCNN \cite{deng2020voxel}) is the total number of grids in a 3D RoI, and $C$ is the grid feature channel. 
The $i^{\rm{th}}$ RoI grid feature of $F^{\rm{raw}}$ and $F^{\rm{pse}}$ are denoted as $F_i^{\rm{raw}}$ and $F_i^{\rm{pse}}$, respectively.
Given a pair of RoI grid features $(F_i^{\rm{raw}}, F_i^{\rm{pse}})$,
we concatenate the $F_i^{\rm{raw}}$ and $F_i^{\rm{pse}}$. Then the result is fed to a fully connected layer and a sigmoid layer, producing a pair of weights $(w_i^{\rm{raw}}, w_i^{\rm{pse}})$ for the pair of grid features, where $w_i^{\rm{raw}}$ and $w_i^{\rm{pse}}$ are all scalars. Finally, we weight $(F_i^{\rm{raw}}, F_i^{\rm{pse}})$ with $(w_i^{\rm{raw}}, w_i^{\rm{pse}})$ to get the fused grid feature $F_i$. Formally, $F_i$ is attained as follow:
\begin{equation}
 (w_i^{\rm{raw}}, w_i^{\rm{pse}}) = \sigma ( {\rm{MLP}}({\rm{CONCAT}}(F_i^{\rm{raw}}, F_i^{\rm{pse}})) )
\end{equation}
\begin{equation}
 F_i = {\rm{MLP}}({\rm{CONCAT}}(w_i^{\rm{raw}} {F_i^{\rm{raw}}}, w_i^{\rm{pse}} {F_i^{\rm{pse}}}))
\end{equation}
 In practice, all pairs of RoI grid features in a batch can be processed in parallel, so our 3D-GAF is efficient. 

\subsection{Synchronized Augmentation}
\label{sec:SynAugment}
Due to the different representations of images and point clouds, it is difficult for multi-modal methods to utilize many data augmentation approaches, such as gt-sampling \cite{yan2018second} and local noising \cite{zheng2020cia}. Insufficient data augmentation greatly limits the performance of many multi-modal methods. Therefore, we present a multi-modal data augmentation method SynAugment to enable our SFD to use all data augmentation approaches tailored to LiDAR-only methods. Concretely, SynAugment consists of two-folds: \textit{manipulate images like point clouds} and \textit{extract image features in 3D Space}.

\textit{Manipulate Images like Point Clouds} \quad
The greatest challenge of multi-modal data augmentation is how to manipulate images like point clouds. Depth completion gives the answer. With depth completion, 2D images can be converted into 3D pseudo clouds. Painting pseudo clouds with RGB, the pseudo clouds carry all information of images. Then we only need to perform data augmentation on pseudo clouds as same as raw clouds, as shown at the top of \fig{fig:method_cpconv_augmentation}.

\textit{Extract Image Features in 3D Space} \quad
Manipulating images like point clouds is not enough for multi-modal data augmentation. Currently, most multi-modal methods need to extract image features on the FOV images. Nevertheless, that will restrict the model from using data augmentation methods (such as gt-sampling and local rotation) that may cause the FOV occlusion issue.
To address this problem, we propose to extract image features in 3D space with 2D images converted to 3D pseudo clouds. In this way, it is unnecessary to consider the occlusion issue because we no longer extract image features on the FOV images. To extract features in 3D space, we can use 3D sparse convolutions. However, there is a more effective method (see Section \ref{sec:CPConv}).

It is noteworthy that \cite{zhang2020multi, wang2021pointaugmenting} can realize multi-modal gt-sampling by performing additional occlusion detection on images, while they are not suitable for more complicated data augmentation, which cannot be simply solved by occlusion detection, such as local noising \cite{zheng2020cia} and SA-DA \cite{zheng2021se}. Some works \cite{vora2020pointpainting, xie2020pi} that project image segmentation scores to raw clouds can also use multi-modal data augmentation, but the image information carried by raw clouds is sparse due to the sparse correspondence between images and point clouds. In our method, the image information of each gt sampler is dense because we can crop complete image information of samplers in pseudo clouds.

\subsection{Color Point Convolution}
\label{sec:CPConv}
\textbf{Definition} \quad
For a frame of pseudo clouds $\mathcal{P}$, we concatenate the RGB ($r$, $g$, $b$) and coordinate ($u$, $v$) of each pixel in the image to its corresponding pseudo point. Therefore, the $i^{\rm{th}}$ pseudo point $p_i$ can be represented as ($x_i$, $y_i$, $z_i$, $r_i$, $g_i$, $b_i$, $u_i$, $v_i$). 

A naive approach to extract features of pseudo clouds is directly voxelizing the pseudo clouds and performing 3D sparse convolutions, while it actually does not fully explore the rich semantic and structural information in pseudo clouds.
PointNet++ \cite{qi2017pointnet++} is a good example for extracting features of points, but it is not suitable for pseudo clouds. \textit{Firstly}, the ball query operation in PointNet++ will bring massive calculations due to the vast amounts of pseudo points. \textit{Secondly}, PointNet++ cannot extract 2D features because the ball query operation does not take 2D neighborhood relationships into account. In light of this, we need a feature extractor that can efficiently extract both 2D semantic features and 3D structural features.

\textit{RoI-aware Neighbor Search on the Image Domain} \quad
Based on the above analysis, we propose a CPConv (Color Point Convolution), which searches neighbors on the \textit{image domain}, as inspired by the voxel query \cite{deng2020voxel} and grid search \cite{fan2021rangedet}. In this way, we can overcome the shortcomings of PointNet++. \textit{Firstly}, a pseudo point can search its neighbors in constant time, making it much faster than the ball query. \textit{Secondly}, neighborhood relationships on the image domain make it possible to extract 2D semantic features.

Unfortunately, we cannot project all pseudo points to current frame image space for neighbor search, because with gt-sampling, pseudo points coming from other frames may cause FOV occlusions.
To this end, we propose an \textit{RoI-aware Neighbor Search}. Concretely, we project pseudo points in each 3D RoI to their original image space separately according to the ($u$, $v$) attribute carried on pseudo points, as shown at the bottom of \fig{fig:method_cpconv_augmentation}. In this way, pseudo points occluded by each other will not become neighbors, so their features will not interfere with each other even if there are heavy occlusions between them on the FOV.

\begin{figure}[t]
	\vspace{-4pt}
	\begin{center}
		\setlength{\fboxrule}{0pt}
		\fbox{\includegraphics[width=0.45\textwidth]{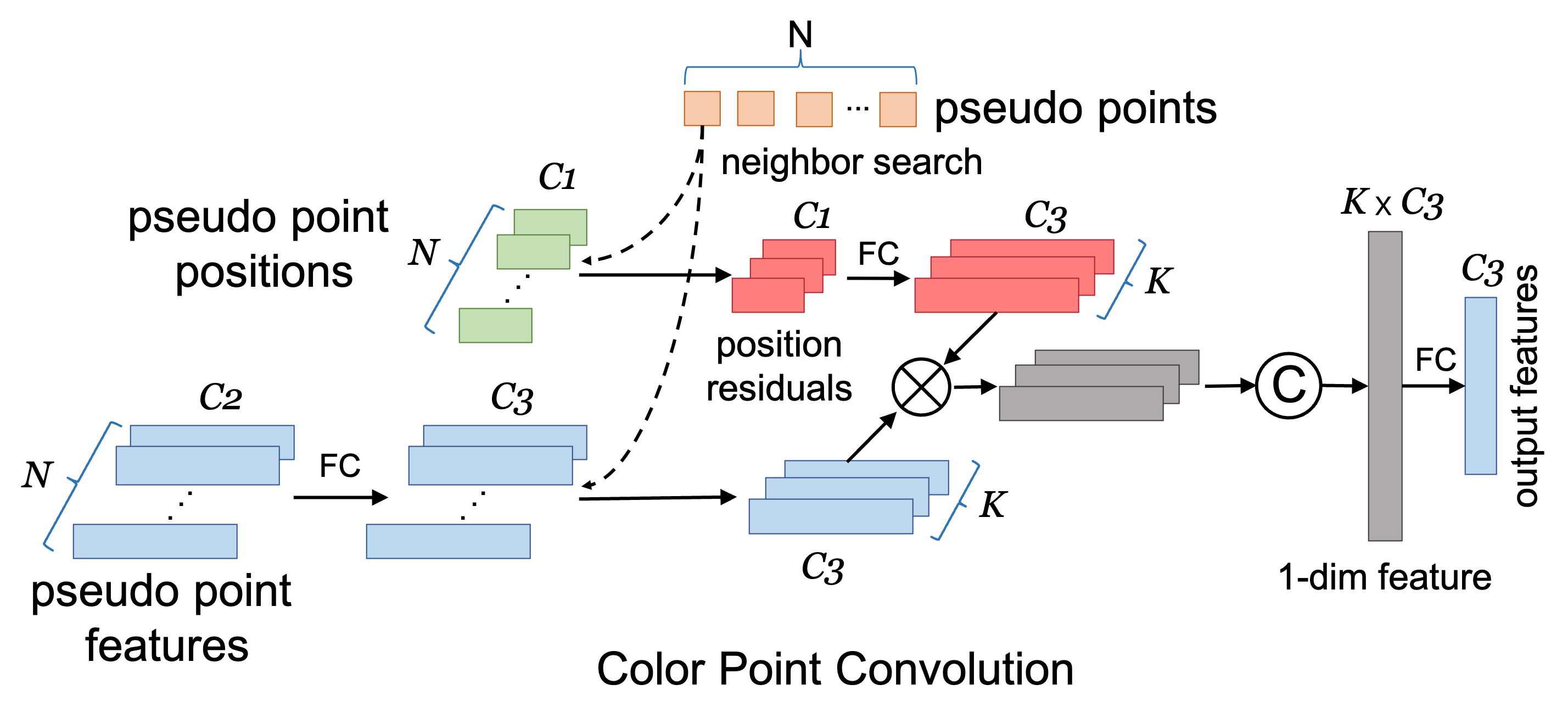}}
	\end{center}
	\captionsetup{font={small}}
	\vspace{-18pt}
	\caption{Illustration of CPConv.}
	\label{fig:method_cpconv}
	\vspace{-8pt}
\end{figure}

\textit{Pseudo Point Features} \quad
For the $i^{\rm{th}}$ pseudo point $p_i$, we denote the feature of $p_i$ as $f_i$ = ($x_i$, $y_i$, $z_i$, $r_i$, $g_i$, $b_i$), which consists of 3D geometric features ($x_i$, $y_i$, $z_i$) and 2D semantic features ($r_i$, $g_i$, $b_i$). As motivated by \cite{deng2020voxel}, we apply a fully connected layer on pseudo point features before performing the neighbor search to reduce the complexity. After the fully connected layer, the feature channel is raised to $C_3$, as shown in \fig{fig:method_cpconv}.

\textit{Position Residuals} \quad
We utilize 3D and 2D position residuals from $p_i$ to its neighbors to make $p_i$'s features aware of local relationships in 3D and 2D space, which is particularly important for extracting both 3D structural features and 2D semantic features of $p_i$. For $p_i$'s $k^{\rm{th}}$ neighbor ${p_i}^{k}$, the position residual between $p_i$ and ${p_i}^{k}$ is represented as ${h_i}^{k}$ = ($x_i-{x_i}^{k}$, $y_i-{y_i}^{k}$, $z_i-{z_i}^{k}$, $u_i-{u_i}^{k}$, $v_i-{v_i}^{k}$, $||p_i - {p_i}^{k}||$), where $||p_i - {p_i}^{k}|| = \sqrt{(x_i-{x_i}^{k})^2 + (y_i-{y_i}^{k})^2 + (z_i-{z_i}^{k})^2}$.

\textit{Feature Aggregation} \quad
For $K$ neighbors of $p_i$, we gather their positions and calculate position residuals. Then we apply a fully connected layer on position residuals, raising their channels to $C_3$ to align with pseudo point features. Given a set of neighbor features $F_i=\{f_i^{k} \in \mathbb{R}^{C_3}, k \in 1, \cdots, K\}$ and a set of neighbor position residuals $H_i=\{h_i^{k} \in \mathbb{R}^{C_3}, k \in 1, \cdots, K\}$, we weight each ${f_i}^k$ with corresponding ${h_i}^k$. The weighted neighbor features are concatenated \cite{fan2021rangedet} instead of max-pooled \cite{deng2020voxel} for maximum information fidelity. Finally, a fully connected layer is applied to map aggregated feature channel back to $C_3$.

\textit{Multi-Level Feature Fusion} \quad
We stack three CPConvs to extract deeper features of pseudo clouds. Considering that high-level features provide a larger receptive field and richer semantic information, while low-level features can supply finer structure information, we concatenate the output of each CPConv to get a more comprehensive and discriminative representation for pseudo clouds.

\subsection{Loss Function} 
\label{sec:Loss}
We follow the RPN loss and RoI head loss of Voxel-RCNN \cite{deng2020voxel}, which are denoted as $\mathcal{L}_{\text{rpn}}$ and $\mathcal{L}_{\text{roi}}$, respectively. 
To prevent gradients from being dominated by a particular \textit{Stream}, we add auxiliary RoI head loss on both \textit{LiDAR Stream} and \textit{Pseudo Stream}, which are denoted as $\mathcal{L}_{{\text{aux}_1}}$ 
and $\mathcal{L}_{{\text{aux}}_2}$, respectively. $\mathcal{L}_{{\text{aux}_1}}$ and $\mathcal{L}_{{\text{aux}_2}}$ are consistent with $\mathcal{L}_{\text{roi}}$, including class confidence loss and regression loss. The depth completion network loss $\mathcal{L}_{\text{depth}}$ follows the definition of \cite{hu2021penet}. Then the total loss is:
\begin{equation}
\mathcal{L} = \mathcal{L}_{\text{rpn}} + \mathcal{L}_{\text{roi}} + \lambda_1 \mathcal{L}_{{\text{aux}}_1} + \lambda_2 \mathcal{L}_{{\text{aux}}_2} + \beta \mathcal{L}_{\text{depth}}
\end{equation}
where ${\lambda}_1$, ${\lambda}_2$ and $\beta$ are the weight of $\mathcal{L}_{{\text{aux}}_1}$, $\mathcal{L}_{{\text{aux}}_2}$ and $\mathcal{L}_{\text{depth}}$ (${\lambda}_1=0.5$, ${\lambda}_2=0.5$, $\beta=1$ by default). More details about the methods that we propose in this paper are provided in the supplementary material.
\section{Experiments}
\label{sec:Experiments}

\subsection{Dataset and Evaluation Metrics}
We evaluate our method on the KITTI 3D and BEV object detection benchmark \cite{geiger2013vision}. The KITTI dataset consists of 7481 training samples and 7518 testing samples in the object detection task. The training data are divided into a \textit{train} set with $3712$ samples and a \textit{val} set with $3769$ samples. 
For experimental studies, we use the \textit{train} set and \textit{val} set for training and evaluating, respectively. 
The results on the \textit{val} set and \textit{test} set are evaluated with the average precision calculated by 40 recall positions.
We also provide results on \textit{val} set with AP calculated by 11 recall positions for a fair comparison with previous works.
For the reason that Waymo and NuScenes datasets have not yet generated depth labels for the depth completion task, we do not conduct experiments on these two datasets.

\subsection{Implementation Details}
\label{sec:details}
The \textit{LiDAR Stream} of SFD is based on Voxel-RCNN \cite{deng2020voxel}.
For the depth completion, we use \cite{hu2021penet}. SFD can also achieve comparable results with \cite{imran2021depth} as our depth completion network. We follow the data augmentation approaches mentioned in \cite{deng2020voxel} (gt-sampling, global rotation, global flipping and global scaling) and \cite{zheng2020cia} (local noising and training with similar class). Although our SFD can be trained end-to-end without the depth completion network pre-trained, we observe that initialization is essential for the performance of 3D detection. Thus, we pre-train the depth completion network on the KITTI dataset and fix the parameters of the depth completion network when training our SFD. 

\subsection{Comparison with State-of-the-Arts}
We compare our SFD with state-of-the-art methods on the KITTI $test$ set by submitting our results to KITTI online test server. As shown in \tab{tab:kitti_test_results}, our method achieves remarkable results. We surpass all state-of-the-art multi-modal methods by a large margin. For LiDAR-only methods, we improve our baseline Voxel-RCNN by $3.14\%$ AP on the moderate metric and outperform published best method SE-SSD \cite{zheng2021se} by $2.22\%$ and $1.07\%$ AP on the moderate and mAP metric, respectively. As of Nov.16, 2021, our method ranks $1^{st}$ on the highly competitive KITTI car 3D detection benchmark. Besides, we provide a comparison on the KITTI $val$ set, as seen in \tab{tab:kitti_val_results}. In BEV detection, SFD is still in the leading position, as shown in \tab{tab:kitti_bev_results}. We improve Voxel-RCNN by $3.02\%$ AP on the moderate metric and achieve comparable results with the state-of-the-art method SE-SSD.
\begin{table}[t]
	\centering
	\footnotesize
	\scalebox{0.88}[0.88]{
		\setlength\tabcolsep{6pt}
		\begin{tabular}{c|c|cccc}
			\hline
			\multicolumn{1}{c|}{ \multirow{2}{*}{Method}} 
			& \multicolumn{1}{c|}{ \multirow{2}{*}{Modality}} 
			& \multicolumn{4}{c}{3D} \\
			\cline{3-6}
			\multicolumn{1}{c|}{} 
			& \multicolumn{1}{c|}{} 
			& \multicolumn{1}{c}{mAP} & \multicolumn{1}{c}{Easy} & \multicolumn{1}{c}{Mod.} & \multicolumn{1}{c}{Hard} \\
			\hline
			\hline
			SECOND~\cite{yan2018second}                  &{LiDAR}       & 73.90 & 83.34  & 72.55  & 65.82 \\
			PointPillars~\cite{lang2019pointpillars}     &{LiDAR}       & 75.29 & 82.58  & 74.31  & 68.99 \\
			Part-$A^2$~\cite{shi2020points}              &{LiDAR}       & 79.94 & 87.81  & 78.49  & 73.51 \\
			SA-SSD~\cite{he2020structure}                &{LiDAR}       & 80.90 & 88.75  & 79.79  & 74.16 \\
			PV-RCNN~\cite{shi2020pv}                     &{LiDAR}       & 82.83 & 90.25  & 81.43  & 76.82 \\
			Voxel-RCNN~\cite{deng2020voxel}              &{LiDAR}       & 83.19 & 90.90  & 81.62  & 77.06 \\
			CT3D~\cite{sheng2021improving}               &{LiDAR}       & 82.25 & 87.83  & 81.77  & 77.16 \\
			Pyramid R-CNN~\cite{mao2021pyramid}          &{LiDAR}       & 82.65 & 88.39  & 82.08  & 77.49 \\
			VoTr-TSD~\cite{mao2021voxel}                 &{LiDAR}       & 83.71 & 89.90  & 82.09  & \textbf{79.14} \\
			SPG~\cite{xu2021spg}                         &{LiDAR}       & 83.84 & 90.50  & 82.13  & 78.90 \\
			SE-SSD~\cite{zheng2021se}                    &{LiDAR}       & 83.73 & 91.49  & 82.54  & 77.15 \\
			\hline
			\hline
			MV3D~\cite{MV3D}                             & {LiDAR+RGB}  & 64.20 & 74.97  & 63.63  & 54.00 \\
			ContFuse~\cite{CONTFUSE}                     & {LiDAR+RGB}  & 71.38 & 83.68  & 68.78  & 61.67 \\
			F-PointNet~\cite{qi2018frustum}              & {LiDAR+RGB}  & 70.86 & 82.19  & 69.79  & 60.59 \\
			AVOD~\cite{AVOD}                             & {LiDAR+RGB}  & 73.52 & 83.07  & 71.76  & 65.73 \\
			PI-RCNN~\cite{xie2020pi}                     & {LiDAR+RGB}  & 76.41 & 84.37  & 74.82  & 70.03 \\
			UberATG-MMF~\cite{liang2019multi}            & {LiDAR+RGB}  & 78.68 & 88.40  & 77.43  & 70.22 \\
			EPNet~\cite{liang2019multi}                  & {LiDAR+RGB}  & 81.23 & 89.81  & 79.28  & 74.59 \\
			3D-CVF~\cite{yoo20203d}                      & {LiDAR+RGB}  & 80.79 & 89.20  & 80.05  & 73.11 \\
			CLOCs PVCas~\cite{pang2020clocs}             & {LiDAR+RGB}  & 82.25 & 88.94  & 80.67  & 77.15 \\
			
			\cline{1-6}
			\bf SFD (ours)                               & {LiDAR+RGB}  & \textbf{84.80} & \textbf{91.73}  &\textbf{84.76}  & 77.92 \\
			\hline
		\end{tabular}
	}
	\vspace{-1mm}
	\caption{Comparison with state-of-the-art methods on the KITTI \textit{test} set for car 3D detection, with average precisions of 40 sampling recall points evaluated on the KITTI server.
	}
	\label{tab:kitti_test_results}
	\vspace{-1mm}
\end{table}
\begin{table}[t]
	\centering
	\footnotesize
	\scalebox{0.9}[0.9]{
		\setlength\tabcolsep{6pt}
		\begin{tabular}{c|ccc|ccc}
			\hline
			\multicolumn{1}{c|}{ \multirow{2}{*}{Method}} 
			& \multicolumn{3}{c|}{3D$_{R11}$} 
			& \multicolumn{3}{c}{3D$_{R40}$}\\
			\cline{2-7}
			\multicolumn{1}{c|}{} 
			& \multicolumn{1}{c}{Easy} & \multicolumn{1}{c}{Mod.} & \multicolumn{1}{c|}{Hard} 
			& \multicolumn{1}{c}{Easy} & \multicolumn{1}{c}{Mod.} & \multicolumn{1}{c}{Hard}  \\
			\hline
			\hline
			PV-RCNN~\cite{shi2020pv}                      & 89.35 & 83.69 & 78.70  & 92.57 & 84.83 & 82.69 \\
			Pyramid-PV~\cite{mao2021pyramid}              & 89.37 & 84.38 & 78.84  & - & - & - \\
			Voxel-RCNN~\cite{deng2020voxel}               & 89.41 & 84.52 & 78.93  & 92.38 & 85.29 & 82.86 \\
			SE-SSD~\cite{zheng2021se}                     & -     & 85.71 & -      & 93.19 & 86.12 & 83.31  \\ 
			\hline
			\hline
			UberATG-MMF~\cite{liang2019multi}             & 88.40 & 77.43 & 70.22 & - & - & - \\ 
			3D-CVF~\cite{yoo20203d}                       & -     & -     & -     & 89.67 & 79.88 & 78.47 \\ 
			EPNet~\cite{huang2020epnet}                   & -     & -     & -     & 92.28 & 82.59 & 80.14 \\
			CLOCs PVCas~\cite{pang2020clocs}              & -     & -     & -     & 92.78 & 85.94 & 83.25 \\
			
			\cline{1-7}
			\bf SFD (ours)                                & \textbf{89.74} & \textbf{87.12} & \textbf{85.20}  & \textbf{95.47} & \textbf{88.56} & \textbf{85.74} \\
			\hline
		\end{tabular}
	}
	\vspace{-1mm}
	\caption{Comparison with state-of-the-art methods on the KITTI \textit{val} set for car 3D detection. The results are evaluated with the average precision calculated by 11 and 40 recall positions.
	}
	\label{tab:kitti_val_results}
	\vspace{-1mm}
\end{table}
\begin{table}[t]
	\centering
	\footnotesize
	\scalebox{0.88}[0.88]{
		\setlength\tabcolsep{6pt}
		\begin{tabular}{c|c|cccc}
			\hline
			\multicolumn{1}{c|}{ \multirow{2}{*}{Method}} 
			& \multicolumn{1}{c|}{ \multirow{2}{*}{Modality}} 
			& \multicolumn{4}{c}{BEV} \\
			\cline{3-6}
			\multicolumn{1}{c|}{} 
			& \multicolumn{1}{c|}{} 
			& \multicolumn{1}{c}{mAP} & \multicolumn{1}{c}{Easy} & \multicolumn{1}{c}{Mod.} & \multicolumn{1}{c}{Hard} \\
			\hline
			\hline
			Voxel-RCNN~\cite{deng2020voxel}              &{LiDAR}       & 89.94 & 94.85 & 88.83 & 86.13 \\
			SA-SSD~\cite{he2020structure}                &{LiDAR}       & 90.67 & 95.03 & 91.03 & 85.96 \\
			SE-SSD~\cite{zheng2021se}                    &{LiDAR}       & 91.41 & \textbf{95.68} & 91.84 & 86.72 \\
			\hline
			\hline
			EPNet~\cite{liang2019multi}                  & {LiDAR+RGB}  & 88.79 & 94.22 & 88.47 & 83.69 \\ 
			3D-CVF~\cite{yoo20203d}                      & {LiDAR+RGB}  & 88.51 & 93.52 & 89.56 & 82.45 \\
			CLOCs PVCas~\cite{pang2020clocs}             & {LiDAR+RGB}  & 89.81 & 93.05 & 89.80 & 86.57 \\
			
			\cline{1-6}
			\bf SFD (ours)                               & {LiDAR+RGB}  & \textbf{91.44} & 95.64 &\textbf{91.85} &\textbf{86.83} \\
			\hline
		\end{tabular}
	}
	\vspace{-1mm}
	\caption{Comparison with state-of-the-art methods on the KITTI \textit{test} set for car BEV detection, with average precisions of 40 sampling recall points evaluated on the KITTI server.
	}
	\label{tab:kitti_bev_results}
	\vspace{-4mm}
\end{table}

\subsection{Ablation Study}
Here we provide extensive experiments to analyze the effectiveness of our method. 
In \tab{tab:ablation_module_stack}, experiment (a) is our baseline modified on Voxel-RCNN \cite{deng2020voxel}. It only uses raw clouds as input. Experiments (b) and (c) are all equipped with our multi-modal data augmentation method SynAugment for a fair comparison with experiment (a), which is equipped with single-modal data augmentation. 

\vspace{-3mm}
\paragraph{Effect of 3D-GAF} 
Experiment (b) in \tab{tab:ablation_module_stack} exploit 3D-GAF to fuse RoI features, making $0.61\%$, $1.10\%$ and $2.32\%$ AP improvement on easy, moderate and hard levels, respectively. To extract pseudo RoI features, we simply voxelize pseudo clouds and perform 3D sparse convolutions.

\vspace{-3mm}
\paragraph{Effect of CPConv}
Experiment (c) in \tab{tab:ablation_module_stack} uses CPConv to extract richer features of pseudo clouds based on experiment (b), yielding a moderate AP of $88.56\%$ with $1.99\%$ AP improvement, manifesting the effectiveness of CPConv.

\begin{table}[h]
	\vspace{1mm}
	\begin{center}
		\scalebox{0.8}[0.8]{
			\setlength\tabcolsep{6pt}
			\begin{tabular}{c|cc|ccc}
				\hline
				\multirow{2}*{Experiment} & \multirow{2}*{3D-GAF} & \multirow{2}*{CPConv} & \multicolumn{3}{c}{$\text{AP}_\text{3D}$} \\
				\cline{4-6}
				&                       &                     & Easy   & Mod. & Hard \\
				\hline
				\hline
				(a)                       &              		  &                     & 92.88  & 85.47    & 82.98 \\
				(b)                       & $\bm{\surd}$ 		  &                     & 93.49  & 86.57    & 85.30 \\
				(c)                       & $\bm{\surd}$ 		  & $\bm{\surd}$        & \textbf{95.47}  & \textbf{88.56}    & \textbf{85.74} \\
				\hline
			\end{tabular}
		}
	\end{center}
	\vspace{-4mm}
	\caption{Effects of different components on the KITTI \textit{val} set. The results are evaluated with the AP calculated by 40 recall positions for car class. ``3D-GAF'' and ``CPConv'' stand for 3D Grid-wise Attentive Fusion and Color Point Convolution, respectively.
	}
	\label{tab:ablation_module_stack}
	\vspace{-4mm}
\end{table}

\vspace{-3mm}
\paragraph{Effect of SynAugment}
Our SynAugment enables our multi-modal framework to utilize the data augmentation approaches tailored only for LiDAR-only methods such as gt-sampling, local nosing and global scaling.
We take off these data augmentation approaches from experiments (a) and (b) in \tab{tab:ablation_module_stack}, resulting in experiments (a) and (b) in \tab{tab:ablation_synaugment}. 
As shown in \tab{tab:ablation_synaugment}, without multi-modal data augmentation, the performance of our method drops drastically, which proves the importance of sufficient data augmentation for multi-modal methods.
\begin{table}[h]
		\begin{center}
		\scalebox{0.8}{
			\begin{tabular}{c|c|ccc}
				\hline
				\multirow{2}*{Experiment} & \multirow{2}*{Data Augmentation} & \multicolumn{3}{c}{$\text{AP}_\text{3D}$} \\
				\cline{3-5}
				& &Easy & Mod. & Hard \\
				\hline
				\hline
				\multirow{2}*{(a)} & Yes & \textbf{92.88}  & \textbf{85.47}    & \textbf{82.98} \\
				& No & 88.55 & 78.49 & 74.42          \\
				
				\hline
				\multirow{2}*{(b)} & Yes & \textbf{93.49}  & \textbf{86.57}    & \textbf{85.30} \\
				& No & 90.88 & 80.31 &77.87           \\
				\hline
			\end{tabular}
		}
		\end{center}
		\vspace{-4mm}
		\caption{Ablation study on SynAugment. The results are evaluated with the AP calculated by 40 recall positions for car class.}
		\label{tab:ablation_synaugment}
		\vspace{-4mm}
\end{table}

\vspace{-3mm}
\paragraph{Ablation Study on 3D Grid-wise Attentive Fusion} 
We conduct an experiment to verify the effectiveness of each part of 3D-GAF, as shown in \tab{tab:ablation_3d-gaf}.
Experiment (a) directly concatenates raw RoI features and pseudo RoI features cropped by 2D RoIs, which we call \textit{2D RoI-wise Concat Fusion}.
Experiment (b) concatenates raw RoI features and pseudo RoI features cropped by 3D RoIs, which we call \textit{3D RoI-wise Concat Fusion}.
Experiment (c) fuses a pair of RoI features in a grid-wise manner based on experiment (b), which we call \textit{3D Grid-wise Concat Fusion}. 
Experiment (d) extends (c) with Attentive Fusion, which is our \textit{3D Grid-wise Attentive Fusion}.
Results show that each part of 3D-GAF can improve our SFD. Moreover, we find that the contribution of Grid-wise Fusion and Attentive Fusion mainly lie on the moderate level and easy level, respectively.
\begin{table}[h]
	\renewcommand\arraystretch{1}
	\small
	\begin{center}
		\scalebox{0.8}[0.8]{
			\setlength\tabcolsep{6pt}
			\begin{tabular}{c|ccc|ccc}
				\hline
				\multirow{2}*{Experiment}  & \multirow{2}*{3D}   & \multirow{2}*{Grid-wise} & \multirow{2}*{Attentive}    & \multicolumn{3}{c}{$\text{AP}_\text{3D}$} \\
				\cline{5-7}
				                           &                     &                          &                             &Easy &Moderate &Hard \\
				\hline
				\hline
				(a)                        &                     &                          &                             &93.08 &85.27 &82.79 \\
				(b)                        &$\bm{\surd}$         &                          &                             &94.83 &87.77 &85.27 \\
				(c)                        &$\bm{\surd}$         &$\bm{\surd}$              &                             &94.84 &88.23 &85.57 \\
				(d)                        &$\bm{\surd}$         &$\bm{\surd}$              & $\bm{\surd}$                &\textbf{95.47} &\textbf{88.56} &\textbf{85.74} \\
				\hline
			\end{tabular}
		}
	\end{center}
	\vspace{-3mm}
	\caption{Ablation study on 3D-GAF.\quad ``3D'': 3D Fusion. ``Grid-wise'': Grid-wise Fusion. ``Attentive'': Attentive Fusion. The results are calculated by 40 recall positions for car class.}
	\label{tab:ablation_3d-gaf}
	\vspace{-4mm}
\end{table}

\paragraph{Cooperating with Different Detectors} 
To validate the universality of our method, we equip different LiDAR-only detectors with our SFD framework. In our experiments, we use the PointRCNN \cite{shi2019pointrcnn}, Part-$A^2$ \cite{shi2020points} and SECOND \cite{yan2018second} implemented by OpenPCDet \cite{openpcdet2020}. \tab{tab:ablation_detectors} suggests that our method can improve different detectors significantly. For the one-stage detector SECOND, we use the same architecture as \textit{Pseudo Stream} (CPConvs with sparse convolutions) to extract features of raw clouds in 3D RoIs. The raw clouds are also painted with RGB to be consistent with pseudo clouds.
\begin{table}[h]
	\small	
	\vspace{-1mm}
	\begin{center}
		\scalebox{0.8}{
			\begin{tabular}{c|c|ccc}
				\hline
				\multirow{2}*{Method}  & \multirow{2}*{with SFD} & \multicolumn{3}{c}{$\text{AP}_\text{3D}$} \\
				\cline{3-5}
				&     &Easy    & Mod. & Hard \\
				\hline
				\hline
				\multirow{3}*{PointRCNN}  & No  &91.40 &82.33 &80.09 \\
				& Yes &\textbf{94.50} &\textbf{85.72} &\textbf{83.29} \\
				&\textit{Improvement} &\textit{+3.10} &\textit{+3.39}  &\textit{+3.20}\\
				\hline
				\multirow{3}*{Part-$A^2$} & No  &91.87 &82.74 &80.42 \\
				& Yes &\textbf{93.17} &\textbf{85.91} &\textbf{83.56} \\
				&\textit{Improvement} &\textit{+1.30} &\textit{+3.17} &\textit{+3.14}\\
				\hline
				\multirow{3}*{SECOND}     & No  &90.31 &81.76 &78.88 \\
				& Yes &\textbf{94.75} &\textbf{87.20} &\textbf{85.07} \\
				&\textit{Improvement} &\textit{+4.44} &\textit{+5.44} &\textit{+6.19}\\
				\hline
			\end{tabular}
		}
	\end{center}
	\vspace{-4mm}
	\caption{Cooperating with different detectors. The average precisions are calculated by 40 recall positions.}
	\label{tab:ablation_detectors}
	\vspace{-4mm}
\end{table}

\vspace{-3mm}
\paragraph{Conditional Analysis}
To figure out in what cases our method improves the baseline most, we evaluate our SFD on different distances and different occlusion degrees. As shown in \tab{tab:ablation_far_occ}, distant and heavily occluded objects are improved most, which verifies our hypothesis that pseudo point clouds are helpful for objects with sparse raw points. 
\begin{table}[h]
	\renewcommand\arraystretch{1}
	\small
	\vspace{-1mm}
	\begin{center}
		\scalebox{0.8}[0.8]{
			\setlength\tabcolsep{6pt}
			\begin{tabular}{c|ccc|ccc}
				\hline
				\multirow{2}*{with SFD} & \multicolumn{3}{c|}{Distance} & \multicolumn{3}{c}{Occlusion} \\
				\cline{2-7}
				                        &0-20m    &20-40m &40m-Inf &0      &1      &2 \\
				\hline
				\hline
				No                     &94.42 &77.05 &15.03 &62.49 &76.79 &57.46 \\
				Yes                    &\textbf{95.28} &\textbf{79.34} &\textbf{21.91} &\textbf{63.46} &\textbf{80.03} &\textbf{62.68} \\
				\hline
			    \textit{Improvement  } &\textit{+0.86}	&\textit{+2.29}	&\textit{+6.88} &\textit{+0.97}	&\textit{+3.24}	&\textit{+5.22} \\
				\hline
			\end{tabular}
		}
	\end{center}
	\vspace{-4mm}
	\caption{Performance on different distances and different occlusion degrees. The results are evaluated with 3D AP calculated by 40 recall positions for car class on the moderate level.}
	\label{tab:ablation_far_occ}
	\vspace{-4mm}
\end{table}

\vspace{-3mm}
\paragraph{Inference Speed}
We test the inference speed of our SFD on an NVIDIA RTX 2080 Ti GPU. With the depth completion network, the speed of SFD is 10.2 FPS. Because SFD is a multi-modal detector, it is inevitably slower than some single-modal methods. However, in multi-modal methods, SFD is actually not slow, as shown in \tab{tab:ablation_speed}.
\begin{table}[h]
	\renewcommand\arraystretch{1}
	\small
	\vspace{-1mm}
	\begin{center}
		\scalebox{0.75}[0.75]{
			\setlength\tabcolsep{6pt}
			\begin{tabular}{c|c|c|c|c}
				\hline
				\multirow{1}*{SFD} & \multirow{1}*{PointPainting \cite{vora2020pointpainting}} & \multirow{1}*{F-PointNet \cite{qi2018frustum}} & \multirow{1}*{EPNet \cite{huang2020epnet}} & \multirow{1}*{3D-CVF \cite{yoo20203d}} \\
				\cline{1-5}
				\hline
				\hline
				10.2 FPS           & 2.5 FPS       & 5.9 FPS     & 10 FPS & 16.7 FPS     \\
				\hline
			\end{tabular}
		}
	\end{center}
	\vspace{-4mm}
	\caption{Inference speed of different multi-modal methods.}
	\label{tab:ablation_speed}
	\vspace{-4mm}
\end{table}

\vspace{-3mm}
\paragraph{Training with Three Classes}
To further validate the effectiveness of our SFD, we train a single model for car, pedestrian and cyclist detection. As seen in \fig{tab:ablation_car_ped_cyc}, SFD can consistently improve Voxel-RCNN.
\begin{table}[h]
	\small			
	\begin{center}
		\scalebox{0.73}{
			\begin{tabular}{c|c|ccc|ccc}
				\hline
				\multirow{2}*{Class}  & \multirow{2}*{with SFD} & \multicolumn{3}{c|}{$\text{AP}_\text{3D}$}  & \multicolumn{3}{c}{$\text{AP}_\text{BEV}$} \\
				\cline{3-8}
				&     &Easy & Mod. & Hard  &Easy  & Mod. & Hard \\
				\hline
				\hline
				\multirow{3}*{Car}         & No  &89.39 &83.83 &78.73 &90.26 &88.35 &87.81 \\
				                           & Yes &\textbf{95.52} &\textbf{88.27} &\textbf{85.57} &\textbf{96.24} &\textbf{92.09} &\textbf{91.32} \\
				&\textit{Improvement} &\textit{+6.13} &\textit{+4.44} &\textit{+6.84} &\textit{+5.98} &\textit{+3.74} &\textit{+3.51} \\
				\hline
				\multirow{3}*{Pedestrian}  & No  &70.55 &62.92 &57.35  &71.62 &64.95 &61.11 \\
				                           & Yes &\textbf{72.94} &\textbf{66.69} &\textbf{61.59}  &\textbf{75.64} &\textbf{69.71} &\textbf{64.75} \\
				&\textit{Improvement} &\textit{+2.39} &\textit{+3.77} &\textit{+4.24} &\textit{+4.02} &\textit{+4.76} &\textit{+3.64} \\
				\hline
				\multirow{3}*{Cyclist}     & No  &90.04 &71.13 &66.67 &91.71 &74.67 &70.02 \\
			   	                           & Yes &\textbf{93.39} &\textbf{72.95} &\textbf{67.26} &\textbf{93.37} &\textbf{75.31} &\textbf{70.80} \\
				&\textit{Improvement} &\textit{+3.35} &\textit{+1.82} &\textit{+0.59} &\textit{+1.66} &\textit{+0.64} &\textit{+0.78} \\
				\hline
			\end{tabular}
		}
	\end{center}
	\vspace{-4mm}
	\caption{Performance of SFD on the KITTI \emph{val} set for car, pedestrian and cyclist with AP calculated by 40 recall positions.}	
	\label{tab:ablation_car_ped_cyc}
	\vspace{-4mm}
\end{table}

\begin{figure*}[t]
	\centering
	\includegraphics[width=0.98\linewidth]{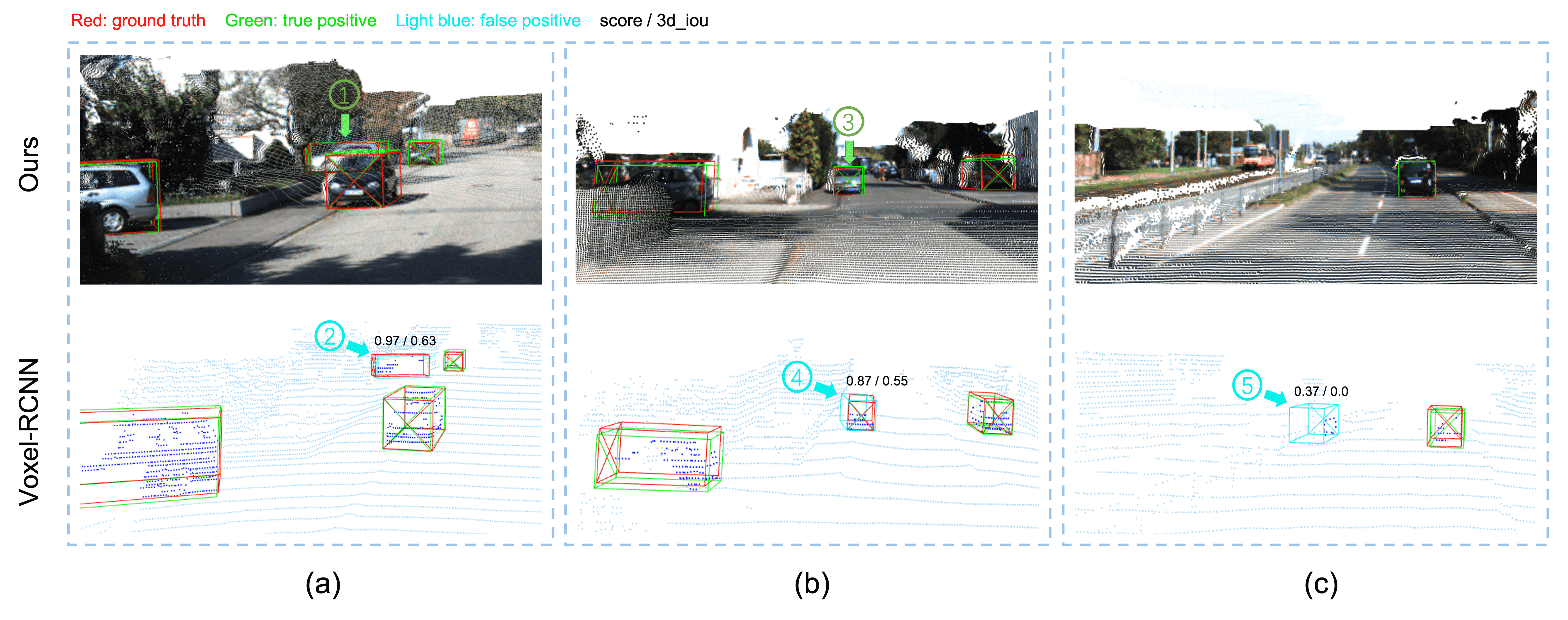}
	\vspace{-8pt}
	\caption{Comparison between our SFD and Voxel-RCNN. For the visualization of SFD and Voxel-RCNN, we use pseudo clouds and raw clouds, respectively. We show ground-truth boxes, true positives and false positives in red, green and light blue, respectively. Green arrows represent that our predictions are more accurate, and light blue arrows represent false positives of Voxel-RCNN.}
	\vspace{-6pt}
	\label{fig:ablation_case_study}
\end{figure*}
\subsection{Qualitative Results and Analysis}
\fig{fig:ablation_case_study} shows the visualization of predictions by our SFD and Voxel-RCNN \cite{deng2020voxel}.
It provides 3 cases corresponding to 3 situations where SFD improves Voxel-RCNN.

\textbf{Occlusion}\quad Occlusion is a challenging problem in the scenario of autonomous driving, as shown in \fig{fig:ablation_case_study}(a).
Object \ding{192} is heavily occluded by the black car in front, making raw clouds on it insufficient (see \ding{193}). Fortunately, pseudo clouds can alleviate this by providing sufficient 3D geometric information and additional 2D semantic information.

\begin{figure}[h]
	\vspace{-8pt}
	\centering
	\includegraphics[width=0.9\linewidth]{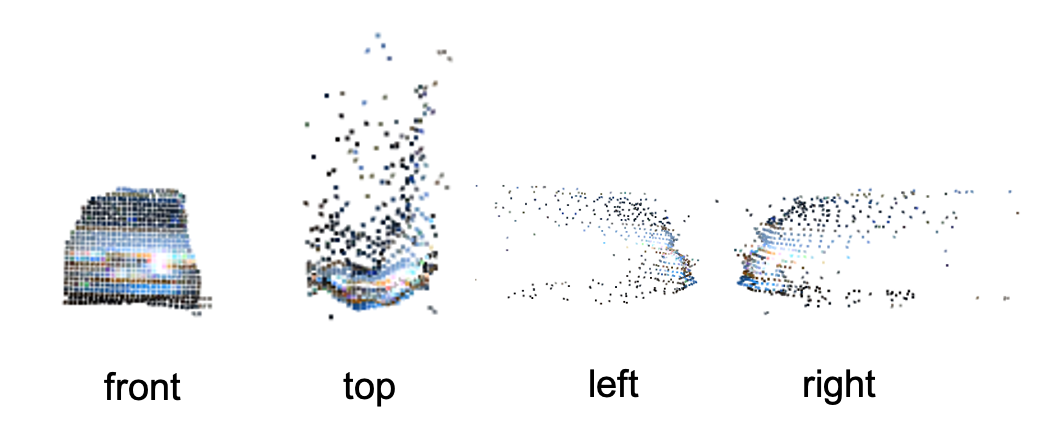}
	\vspace{-10pt}
	\caption{Different views of object \ding{194} in \fig{fig:ablation_case_study}(b).}
	\vspace{-15pt}
	\label{fig:ablation_full_view}
\end{figure}

\textbf{Long Distance}\quad \fig{fig:ablation_case_study}(b) shows another common scene. 
Due to the limited resolution of LiDAR, faraway objects are with much fewer points. It is difficult to predict a precise box for objects (such as \ding{195}) with sparse raw clouds. However, pseudo clouds on the object are richer. \fig{fig:ablation_full_view} shows different views of pseudo clouds on \ding{194}, demonstrating that pseudo clouds are qualified to provide supplementary information for raw clouds.

\textbf{Background similar to Foreground}\quad Dense pseudo clouds not only benefit locating foreground but also help to distinguish the background from the foreground. Some background raw clouds are very similar to the foreground because of the sparsity of raw clouds, which may confuse detectors and cause a lot of false positives. As seen in \fig{fig:ablation_case_study}(c), Voxel-RCNN mistakes the fence for a car because raw clouds on the fence and car are similar. Nevertheless, pseudo clouds on them are very different, which helps our SFD to distinguish them. 

\section{Conclusion}
\label{sec:conclusion}
In this paper, we propose a novel multi-modal framework SFD for high quality 3D detection. We design a new RoI fusion strategy 3D-GAF to fuse raw clouds and pseudo clouds in a more fine-grained manner. With the proposed SynAugment, our SFD can use data augmentation methods tailored to LiDAR-only methods. Besides, we design a CPConv to effectively and efficiently extract features of pseudo clouds. Experimental results demonstrate that our approach can significantly improve detection accuracy.
\\
\\
\noindent\textbf{Acknowledgments} This work was supported in part by The National Key Research and Development Program of China (Grant Nos: 2018AAA0101400), in part by The National Nature Science Foundation of China (Grant Nos: 62036009, U1909203, 61936006, 61973271), in part by Innovation Capability Support Program of Shaanxi (Program No. 2021TD-05).
\clearpage
{\small
	\bibliographystyle{ieee_fullname}
	\bibliography{egbib}
}
\end{document}